\documentclass[11pt]{article}

\usepackage{amsmath}

\usepackage{tikz}
\usepackage[utf8]{inputenc} 
\usepackage[T1]{fontenc}    
\usepackage{hyperref}       
\usepackage{url}            
\usepackage{booktabs}       
\usepackage{amsfonts}       
\usepackage{nicefrac}       
\usepackage{microtype}      
\usepackage{cleveref}       
\usepackage{lipsum}         
\usepackage{graphicx}
\usepackage{natbib}
\usepackage{doi}
\usepackage{enumerate}
\usepackage{mathtools}
\usepackage{amsthm}
\usepackage{subfigure}
\usepackage{gensymb}
\usepackage[margin=1in]{geometry}
\usepackage{algorithm}
\usepackage{algorithmic}
\usepackage{enumitem}

\usepackage{authblk}
 \newcommand{\ignore}[1]{}
 
\usepackage{amssymb}
\usepackage{amsmath}
\usepackage{ bbold }

\newtheorem{mydef}{Definition}
\newtheorem{mythm}{Theorem}

\newtheorem{mylem}{Lemma}
\newtheorem{mycor}{Corollary}

\newcommand{\cat}{\mathcal{C}}

\newcommand{\ob}{\mathrm{Ob}}
\renewcommand{\hom}{\mathrm{Hom}}
\newcommand{\id}{\mathrm{id}}
\newcommand{\op}{\mathrm{op}}

\newcommand{\set}{\textbf{Set}}
\newcommand{\cop}{{\cat^\mathrm{op}}}

\newcommand{\homc}{\hom_\cat}

\newcommand{\llim}{\underleftarrow{\lim}~}
\newcommand{\illim}{``\underleftarrow{\lim}"}
\newcommand{\rlim}{\underrightarrow{\lim}~}
\newcommand{\irlim}{``\underrightarrow{\lim}"}
\newcommand{\catw}{{\mathcal{C}^\wedge}}
\newcommand{\catv}{{\mathcal{C}^\vee}}

\newcommand{\iop}{{I^{\op}}}

\newcommand{\hc}{{h_\cat}}
\newcommand{\kc}{{k_\cat}}

\newcommand{\dec}{{Deconcept}}
\title{
Succinct Representations for Concepts
}

\author[1,2,3]{Yang Yuan}

\affil[1]{\footnotesize IIIS, Tsinghua University}
\affil[2]{\footnotesize Shanghai Artificial Intelligence Laboratory}
\affil[3]{\footnotesize Shanghai Qi Zhi Institute}

\let\svthefootnote\thefootnote
\newcommand\freefootnote[1]{%
	\let\thefootnote\relax%
	\footnotetext{#1}%
	\let\thefootnote\svthefootnote%
}
\begin{document}
\maketitle

\begin{abstract}
Foundation models like chatGPT have demonstrated remarkable performance on various tasks. However, for many questions, they may produce false answers that look accurate. How do we train the model to precisely understand the concepts?  In this paper, we introduce succinct representations of concepts based on category theory. Such representation yields concept-wise invariance properties under various tasks, resulting a new learning algorithm that can provably and accurately learn complex concepts or fix misconceptions. Moreover, by recursively expanding the succinct representations, one can generate a hierarchical decomposition, and manually verify the concept by individually examining each part inside the decomposition.
\end{abstract}

\newpage

\section{Introduction}
The existing foundation models like chatGPT are powerful, but with a significant challenge:  
they may generate hard-to-notice false statements if they do not know the correct answer. 
Ideally, the model should precisely understand the concepts, 
and honestly state what it knows and what it does not know, enabling users to trust its responses with confidence. 

One way to teach models a concept is learning through different tasks. For example, if a model can draw different images of dogs, accurately differentiate between dogs and cats, correctly identify that dogs have four legs and two ears, and competently perform various other dog-related tasks, it is very likely that the model understands what a dog is. However, this approach may not be exhaustive, as some important cases may not be covered. 
Theoretically, we can conclude that the model has perfectly understood the concept, only if the model passes \emph{all possible} tasks, which is usually computationally infeasible to learn or evaluate.

In this paper, we show that it is possible to generate succinct representations of a given concept. By doing so, one effectively reduces the problem of passing all possible tasks to finding the correct representation. Such reduction keeps the functionality of the concept invariant, in the sense that replacing a concept with its representation will generate equivalent outputs for all possible tasks. Based on this invariance property, we introduce a new learning algorithm
that can provably and accurately learn complex concepts or fix misconceptions, given enough sampled tasks.

The verification of a given concept can also be significantly simplified by the reduction from evaluation of all possible tasks to evaluation of the representation. However, there is no free lunch. The representation is succinct, but for accurate verification, one has to recursively evaluate the sub-concepts inside the representation. 
The recursive expansion of such representation has finite depth, because not all concepts have succinct representations. For example, what is ``love''? What is ``time''? These basic concepts cannot be further decomposed. 
In other words, we decompose the challenging task of verifying a complex concept, into verifying a hierarchical decomposition with basic concepts as the leaf nodes that also require verification.

However, it is worth pointing out that the foundation models, like chatGPT, have accurate understanding of the relative simple concepts due to their pretraining on large-scale data sets, as tested with billions of tasks. 
Therefore, the learning or verification process of a given concept may stop at any level of the hierarchical decomposition, when the remaining parts seem obviously correct.

Formally, what is a concept? Given a category\footnote{Readers not familiar with category theory may check Section~\ref{sec:prelim} for a basic introduction.} $\cat$, it seems natural to treat all the objects in $\cat$ as concepts. For example, when $\cat$ is the category of natural language, both ``wall'' and ``clock'' are basic concepts, and their combination ``wall clock'' is also a concept. However, not all concepts are representable, or equivalently, have a  formal name in $\cat$. 
By contrast, the category of presheaves is much larger than $\cat$, and the presheaves can capture the unrepresentable concepts. Specifically, presheaves can be seen as a natural extension of objects, as every object $X$ corresponds to a presheaf $\hom_\cat(\cdot, X)$ (or $\hom_\cat(X,\cdot)$), but the reverse is not true. 
Therefore, in this paper, a concept is a presheaf.

Our concept reduction adopts the categorical perspective, where two things are considered equivalent if they are isomorphic. Indeed, consider the case of ``wall clock'', which can be represented as ``clock that can be put on the wall''. Since ``buying a wall clock'' is equivalent to (or isomorphic to) ``buying a clock that can be put on the wall'',  this representation passes the task of being used in the sentence ``buying a \rule{0.5cm}{0.15mm}''.

If isomorphism implies equivalence, 
why should we even care about the tasks? 
It seems that
for a given concept $A$,  it suffices to find any $X\simeq A$ as the representation of $A$. However, if $X$ is not a decomposition, knowing the equivalence between $X$ and $A$ does not provide sufficient information of the model's knowledge of $A$. Moreover, even if $X\simeq A$ is a decomposition,
the sub-concepts in $X$ have to perfectly replace $A$ in all possible tasks, 
to make the hierarchical learning or verification of $A$ feasible. Hence, the task-wise isomorphisms are necessary in addition to the object-wise isomorphism.

In order to characterize these task-wise isomorphisms, we have to first define the set of all possible tasks. Categorically, 
we care about two kinds of tasks: morphisms and functors.
For a given concept $A\in \cat$, 
the morphisms are $\hom_\cat(\cdot, A)$ or $\hom_\cat(A,\cdot)$ that captures the relationship between $A$ and another object in $\cat$, while a functor $F:\cat\rightarrow \cat'$ computes $F(A)\in \cat'$ as the output. 

A succinct representation for a concept $A$ is valid, only if 
for any task, replacing $A$ with the representation generates an equivalent result. 
As we will show,
there are two kinds of representations that meet this criterion, 
denoted as  $\irlim\alpha$ and $\illim\beta$, for $\alpha: I\rightarrow \cat$ and $\beta:I^\op\rightarrow \cat$. 
While the formal definitions of the representations are difficult to comprehend, we notice that they can be informally treated as the ``definition'' and ``induction'' of the concepts, the two handy tools that people use everyday. 
The definition or induction for a given concept are not necessarily unique, just like the limits in this paper.

Within category theory, the equivalence of limits and concept representations seems nothing more than 
a surprising tautology. However, we focus on the learning scenario, which makes the story completely different. Specifically, our goal is to learn a category $\cat$ with a foundation model $f:\cat \rightarrow \catw$ (or $\catv$), see \citet{yuan2022power}. 
In this setting, the morphisms, functors and limits are not known, so $f$ is not necessarily accurate. We assume the existence of a limit extractor $g$, which outputs the limit $g(f,A)$ for concept $A$. 
Such extractor is already included in foundation models like chatGPT. If you use the prompt like ``Please define $A$'', or ``Can you provide typical examples of $A$?'', you will get high quality (but not necessarily accurate) description correspond to a projective or inductive limit.

Since $f$ is not accurate, the invariance properties existed in $\cat$ become the ideal training goals and verification criteria for $f$. Our main theorem says, $f$ precisely understands a concept $A$, if the invariance properties hold for  the hierarchical decomposition of $A$ obtained by recursively expanding the succinct representations of (sub-)concepts. This characterization immediately leads to Algorithm~\ref{alg:learning}, Algorithm~\ref{alg:verify-1} and Algorithm~\ref{alg:verify-2}, which provably learn and verify the concepts for $f$.

Our main contributions in this paper include:
\begin{itemize}[parsep=0.1cm, topsep=0.1cm]
\item Draw connections between concept learning and limits from category theory. 
	\item Introduce new  algorithms that can provably and accurately learn or verify concepts. 
	\item Reduce concept verification to evaluating the corresponding hierarchical concept decomposition, which becomes feasible for human beings when the leaf nodes are known to be accurate.  
\end{itemize}

\section{Preliminaries}
\label{sec:prelim}

Category theory is used in almost all areas of mathematics. Here we only introduce the necessary notions for understanding the results of our paper. Curious readers may check 
\citet{mac2013categories,riehl2017category, adamek1990abstract} for a more comprehensive introduction. 

\subsection{Category basics}
A category $\cat$ has a set of objects $\ob(\cat)$, and 
a set of morphisms $\hom_\cat(X,Y)$ from $X$ to $Y$ 
for every $X, Y\in \ob(\cat)$.
Given $f\in \hom_\cat(X,Y), g\in \hom_\cat(Y,Z)$, we define their composition as $g\circ f\in  \hom_\cat(X,Z)$. Notice that $\circ$ is associative, i.e.,  $(h\circ g) \circ f=h\circ (g\circ f)$. 
For every $X\in \ob(\cat)$, there exists an unique identity morphism $\id_X \in \hom_\cat(X,X)$. 
A morphism $f: X\rightarrow Y$   is an isomorphism if there exists $g: X\leftarrow Y$ such that $f\circ g=\id_Y$ and $g\circ f=\id_X$. 
In this case, we say $X$ and $Y$ are isomorphic and write $X\simeq Y$. 

We consider a universe $\mathcal{U}$\footnote{Check \citet{kashiwara2006categories} for the related definitions.}. 
A category is a $\mathcal{U}$-category, if 
$\hom_\cat(X,Y)$ is $\mathcal{U}$-small for any $X, Y\in \ob(\cat)$. 
A $\mathcal{U}$-category category $\cat$ is $\mathcal{U}$-small if $\ob(\cat)$ is $\mathcal{U}$-small. A category $\cat$ is $\mathcal{U}$-big, if it is not a $\mathcal{U}$-category. For simplicity, below we may not explicitly mention the universe $\mathcal{U}$, and simply write that $\cat$ is a category, a small category, or a big category.

Given a category $\cat$, we define its opposite $\cat^{\op}$ by setting $\ob(\cat^{\op})=\ob(\cat)$ and  $\hom_\cop(X,Y)=\hom_\cat(Y,X)$. Moreover, given $f\in \hom_\cop(X,Y), g\in \hom_\cop(Y,Z)$, the new composition is $g
\mathrel{\overset{\makebox[0pt]{\mbox{\normalfont\tiny\sffamily op}}}{\circ}}  f= f\circ g\in  \hom_\cop(X,Z)$. 

We define $\set$ to be the category of sets, 
where the objects are sets, and $\hom_\set(X,Y)$ is the set of all functions with domain $X$ and codomain $Y$. 
Notice that we ignore the subtleties about the universe for better presentation, so here just assume that $\set$ does not contain strange objects like a set containing all sets.

Functor is like a function between two categories. Given two categories $\cat, \cat'$, a functor $F:\cat\rightarrow \cat'$ 
maps objects from $\cat$ to $\cat'$ with $F:\ob(\cat)\rightarrow \ob(\cat')$ and
morphisms from $\cat$ to $\cat'$ with $F: \hom_\cat(X,Y)\rightarrow \hom_{\cat'}(F(X),F(Y))$ for all $X,Y\in \cat$, so that $F$ preserves identity and composition. Formally, we have $F(\id_X)=\id_{F(X)}$ for all $X\in \cat$, and 
$F(g\circ f)=F(g)\circ F(f)$ for all $f:X\rightarrow Y, g: Y\rightarrow Z$.  
A functor $F:\cat\rightarrow \cat'$ is an isomorphism of categories if there exists $G:\cat'\rightarrow \cat$ such that $G\circ F(X)=X$ and $F\circ G(Y)=Y$ for all $X\in \cat, Y\in \cat'$, and similarly for the morphisms. In this case, we say $\cat$ and $\cat'$ are isomorphic and  write $\cat\simeq\cat'$. 

The morphisms of functors, also called the natural transformation, is the way to transform the functors while preserving the structure. Given two categories $\cat, \cat'$, and two functors $F_1, F_2$ from $\cat$ to $\cat'$. A morphism of functors $\theta:F_1\rightarrow F_2$ has a morphism $\theta_X: F_1(X)\rightarrow F_2(X)$ for all $X\in \cat$ such that for all $f:X\rightarrow Y\in \hom_\cat(X,Y)$, we have $\theta_Y(F_1(f)(F_1(X))) =
F_2(f)(\theta_X(F_1(X)))$. 
We write $F_1\simeq F_2$ if there exists an isomorphism between $F_1$ and $F_2$. 

A presheaf is a functor from $\cop$ to $\set$, 
and $\catw$ is the category of presheaves. 
Similarly, 
a functor from $\cop$ to $\set^\op$ is called a $\set^\op$-valued presheaf, and $\catv$ is the category of $\set^\op$-valued presheaves. 
In this paper we do not make the differentiation, and name both kinds of functors as presheaves, and $\catw, \catv$ as the categories of presheaves. 
Moreover, define $\hc(X)\triangleq \homc(\cdot, X)\in \catw$, and $\kc(X)\triangleq \homc(X, \cdot)\in \catv$.  
The following lemma is fundamental.
\begin{mylem}[Yoneda lemma] 
	\label{lem:yoneda}	
Given $X\in \cat$ we have, 
\begin{enumerate}
		\item[(1)]
	For $A\in \cat^{\wedge}$, $\hom_{\cat^\wedge}(h_\cat(X), A)\simeq
A(X). $
\item[(2)] 
	For $B\in \catv$, $\hom_{\catv}(B, \kc(X))\simeq
B(X). $
\end{enumerate}	
\end{mylem}

Yoneda lemma says $\hc(X)$ and $\kc(X)$ capture all the information of $X$. In this paper, when there is no confusion, we may ignore $\hc(\cdot)$ and $\kc(\cdot)$, and simply write $X$ to denote $\hc(X)\in \catw$ or $\kc(X)\in \catv$. Therefore, Yoneda lemma says, 
$\hom_{\cat^\wedge}(X, A)\simeq
A(X)$ and $\hom_{\catv}(B, X)\simeq
B(X).$

A functor $F$ from $\cop$ to $\set$ (or $\cat$ to $\set$) is representable if there is an isomorphism between $\hc(X)$ (or $\kc(X)$) and $F$ for some $X\in \cat$. Such $X$ is called a representative of $F$. 

Let $L:\cat\rightarrow \cat', R:\cat'\rightarrow \cat$ be two functors. The pair $(L,R)$ is a pair of adjoint functors, or $L$ is a left adjoint functor to $R$, or $R$ is a right adjoint functor to $L$, if there exists an isomorphism from $\cat^\op\times \cat'$ to $\set$: 
$
\hom_{\cat'}(L(\cdot), \cdot)\simeq \hom_\cat(\cdot, R(\cdot)).
$

\subsection{Limits}
A diagram of shape $I$ in a category $\cat$ is a functor $F:I\rightarrow \cat$, which selects objects in $\cat$ correspond to $\ob(I)$, that preserve the morphisms in $I$. 
Given a functor $\beta: I^\op \rightarrow \set$, define its projective limit as
$\llim \beta\triangleq \hom_{I^\wedge}(\text{pt}_{I^\wedge}, \beta)$, where $\text{pt}_{I^\wedge}(i)=\{\text{pt}\}$ for every $i\in I$, and $\{\text{pt}\}$ is the single point set. 
In other words, $\llim\beta$ denotes the set of all natural transformations between $\text{pt}_{I^\wedge}$ and $\beta$. 
Based on this definition for diagrams in  $\set$, we have the general definition of the limits.

\begin{mydef}[Projective and inductive limits]
\label{def:limits}
Given $\alpha: I\rightarrow \cat, \beta: \iop\rightarrow \cat$ with small $I$, 
the inductive limit $\rlim \alpha \in \catv$ and projective limit $\llim\beta\in \catw$ are defined as:
\begin{enumerate}
	\item[(1)] $\rlim\alpha: X\mapsto \llim \homc(\alpha, X)$
	\item[(2)] $\llim \beta: X\mapsto \llim \homc(X, \beta)$	
\end{enumerate}
Here $\homc(\alpha, X)$ is a functor  that maps $i\in I$ to $\homc(\alpha(i), X)$.
Therefore, $\llim \homc(\alpha, X)$ is a  well defined  limit for a diagram in $\set$.
Same argument holds for $ \llim \homc(X, \beta)$.
\end{mydef}

\begin{figure}[tbh]
	\centering
	\addtocounter{figure}{-1}
	\subfigure{
		\begin{minipage}[t]{0.4\linewidth}	
			\begin{center}
				\tikzset{every picture/.style={line width=0.75pt}} 

\begin{tikzpicture}[x=0.75pt,y=0.75pt,yscale=-1,xscale=1]

\draw [color={rgb, 255:red, 255; green, 0; blue, 0 }  ,draw opacity=1 ]   (204.4,194.33) -- (243.4,194.47) ;
\draw [shift={(245.4,194.48)}, rotate = 180.2] [color={rgb, 255:red, 255; green, 0; blue, 0 }  ,draw opacity=1 ][line width=0.75]    (10.93,-3.29) .. controls (6.95,-1.4) and (3.31,-0.3) .. (0,0) .. controls (3.31,0.3) and (6.95,1.4) .. (10.93,3.29)   ;
\draw [color={rgb, 255:red, 255; green, 0; blue, 0 }  ,draw opacity=1 ][fill={rgb, 255:red, 255; green, 0; blue, 0 }  ,fill opacity=1 ]   (263.4,139.48) -- (263.4,178.48) ;
\draw [shift={(263.4,180.48)}, rotate = 270] [color={rgb, 255:red, 255; green, 0; blue, 0 }  ,draw opacity=1 ][line width=0.75]    (10.93,-3.29) .. controls (6.95,-1.4) and (3.31,-0.3) .. (0,0) .. controls (3.31,0.3) and (6.95,1.4) .. (10.93,3.29)   ;
\draw [color={rgb, 255:red, 0; green, 0; blue, 255 }  ,draw opacity=1 ]   (225.07,127) -- (249.4,127.44) ;
\draw [shift={(251.4,127.48)}, rotate = 181.04] [color={rgb, 255:red, 0; green, 0; blue, 255 }  ,draw opacity=1 ][line width=0.75]    (10.93,-3.29) .. controls (6.95,-1.4) and (3.31,-0.3) .. (0,0) .. controls (3.31,0.3) and (6.95,1.4) .. (10.93,3.29)   ;
\draw [color={rgb, 255:red, 0; green, 0; blue, 255 }  ,draw opacity=1 ]   (196.4,139.67) -- (196.4,180.48) ;
\draw [shift={(196.4,182.48)}, rotate = 270] [color={rgb, 255:red, 0; green, 0; blue, 255 }  ,draw opacity=1 ][line width=0.75]    (10.93,-3.29) .. controls (6.95,-1.4) and (3.31,-0.3) .. (0,0) .. controls (3.31,0.3) and (6.95,1.4) .. (10.93,3.29)   ;
\draw [color={rgb, 255:red, 0; green, 0; blue, 255 }  ,draw opacity=1 ]   (116.1,143.75) -- (140.16,179.16) ;
\draw [shift={(141.28,180.82)}, rotate = 235.81] [color={rgb, 255:red, 0; green, 0; blue, 255 }  ,draw opacity=1 ][line width=0.75]    (10.93,-3.29) .. controls (6.95,-1.4) and (3.31,-0.3) .. (0,0) .. controls (3.31,0.3) and (6.95,1.4) .. (10.93,3.29)   ;
\draw [color={rgb, 255:red, 0; green, 0; blue, 255 }  ,draw opacity=1 ]   (89.6,143.75) -- (66.72,181.11) ;
\draw [shift={(65.68,182.82)}, rotate = 301.48] [color={rgb, 255:red, 0; green, 0; blue, 255 }  ,draw opacity=1 ][line width=0.75]    (10.93,-3.29) .. controls (6.95,-1.4) and (3.31,-0.3) .. (0,0) .. controls (3.31,0.3) and (6.95,1.4) .. (10.93,3.29)   ;
\draw [color={rgb, 255:red, 0; green, 0; blue, 0 }  ,draw opacity=0.4 ]   (96.1,163.75) -- (84.56,179.65) ;
\draw [shift={(83.38,181.27)}, rotate = 305.98] [color={rgb, 255:red, 0; green, 0; blue, 0 }  ,draw opacity=0.4 ][line width=0.75]    (10.93,-3.29) .. controls (6.95,-1.4) and (3.31,-0.3) .. (0,0) .. controls (3.31,0.3) and (6.95,1.4) .. (10.93,3.29)   ;
\draw [color={rgb, 255:red, 0; green, 0; blue, 0 }  ,draw opacity=0.4 ]   (114.1,165.75) -- (121.82,178.07) ;
\draw [shift={(122.88,179.77)}, rotate = 237.94] [color={rgb, 255:red, 0; green, 0; blue, 0 }  ,draw opacity=0.4 ][line width=0.75]    (10.93,-3.29) .. controls (6.95,-1.4) and (3.31,-0.3) .. (0,0) .. controls (3.31,0.3) and (6.95,1.4) .. (10.93,3.29)   ;

\draw (189,188.33) node [anchor=north west][inner sep=0.75pt]  [color={rgb, 255:red, 255; green, 0; blue, 0 }  ,opacity=1 ] [align=left] {X};
\draw (256.67,120.33) node [anchor=north west][inner sep=0.75pt]  [color={rgb, 255:red, 255; green, 0; blue, 0 }  ,opacity=1 ] [align=left] {Y};
\draw (251,187.33) node [anchor=north west][inner sep=0.75pt]  [font=\normalsize,color={rgb, 255:red, 255; green, 0; blue, 0 }  ,opacity=1 ] [align=left] {$\displaystyle \{pt\}$};
\draw (221.33,176) node [anchor=north west][inner sep=0.75pt]  [color={rgb, 255:red, 255; green, 0; blue, 0 }  ,opacity=1 ] [align=left] {f};
\draw (274.33,149) node [anchor=north west][inner sep=0.75pt]  [color={rgb, 255:red, 255; green, 0; blue, 0 }  ,opacity=1 ] [align=left] {g};
\draw (175,119.67) node [anchor=north west][inner sep=0.75pt]  [color={rgb, 255:red, 0; green, 0; blue, 255 }  ,opacity=1 ] [align=left] {X$\displaystyle \times _{*}$Y};
\draw (83,127) node [anchor=north west][inner sep=0.75pt]  [color={rgb, 255:red, 0; green, 0; blue, 255 }  ,opacity=1 ] [align=left] {X$\displaystyle \times $Y};
\draw (86,148) node [anchor=north west][inner sep=0.75pt]  [color={rgb, 255:red, 0; green, 0; blue, 0 }  ,opacity=0.4 ] [align=left] {(x, y)};
\draw (56,186) node [anchor=north west][inner sep=0.75pt]  [color={rgb, 255:red, 255; green, 0; blue, 0 }  ,opacity=1 ] [align=left] {X};
\draw (140,186) node [anchor=north west][inner sep=0.75pt]  [color={rgb, 255:red, 255; green, 0; blue, 0 }  ,opacity=1 ] [align=left] {Y};
\draw (73,181.5) node [anchor=north west][inner sep=0.75pt]  [color={rgb, 255:red, 0; green, 0; blue, 0 }  ,opacity=0.4 ] [align=left] {x};
\draw (124,181.5) node [anchor=north west][inner sep=0.75pt]  [color={rgb, 255:red, 0; green, 0; blue, 0 }  ,opacity=0.4 ] [align=left] {y};

\end{tikzpicture}
			\end{center}
			\caption{Projective limit: $X\times Y$}
			\label{fig:product1}
		\end{minipage}
	}
	\subfigure{
		\begin{minipage}[t]{0.4\linewidth}	
			\begin{center}
				\tikzset{every picture/.style={line width=0.75pt}} 

\begin{tikzpicture}[x=0.75pt,y=0.75pt,yscale=-1,xscale=1]
	
	\draw [color={rgb, 255:red, 255; green, 0; blue, 0 }  ,draw opacity=1 ]   (138.07,177.67) -- (180.07,178.3) ;
	\draw [shift={(182.07,178.33)}, rotate = 180.87] [color={rgb, 255:red, 255; green, 0; blue, 0 }  ,draw opacity=1 ][line width=0.75]    (10.93,-3.29) .. controls (6.95,-1.4) and (3.31,-0.3) .. (0,0) .. controls (3.31,0.3) and (6.95,1.4) .. (10.93,3.29)   ;
	\draw [color={rgb, 255:red, 255; green, 0; blue, 0 }  ,draw opacity=1 ]   (138.73,186.33) -- (180.73,186.97) ;
	\draw [shift={(182.73,187)}, rotate = 180.87] [color={rgb, 255:red, 255; green, 0; blue, 0 }  ,draw opacity=1 ][line width=0.75]    (10.93,-3.29) .. controls (6.95,-1.4) and (3.31,-0.3) .. (0,0) .. controls (3.31,0.3) and (6.95,1.4) .. (10.93,3.29)   ;
	\draw [color={rgb, 255:red, 0; green, 0; blue, 255 }  ,draw opacity=1 ]   (127.4,147) -- (127.7,166.33) ;
	\draw [shift={(127.73,168.33)}, rotate = 269.1] [color={rgb, 255:red, 0; green, 0; blue, 255 }  ,draw opacity=1 ][line width=0.75]    (10.93,-3.29) .. controls (6.95,-1.4) and (3.31,-0.3) .. (0,0) .. controls (3.31,0.3) and (6.95,1.4) .. (10.93,3.29)   ;
	
	\draw (122,176.33) node [anchor=north west][inner sep=0.75pt]  [color={rgb, 255:red, 255; green, 0; blue, 0 }  ,opacity=1 ] [align=left] {X};
	\draw (187.07,177.33) node [anchor=north west][inner sep=0.75pt]  [color={rgb, 255:red, 255; green, 0; blue, 0 }  ,opacity=1 ] [align=left] {Y};
	\draw (152,159.67) node [anchor=north west][inner sep=0.75pt]  [color={rgb, 255:red, 255; green, 0; blue, 0 }  ,opacity=1 ] [align=left] {f};
	\draw (150.67,192.33) node [anchor=north west][inner sep=0.75pt]  [color={rgb, 255:red, 255; green, 0; blue, 0 }  ,opacity=1 ] [align=left] {g};
	\draw (111.33,123) node [anchor=north west][inner sep=0.75pt]  [color={rgb, 255:red, 0; green, 0; blue, 255 }  ,opacity=1 ] [align=left] {$\displaystyle f^{-1}$b$\displaystyle \in X$};

\end{tikzpicture}
			\end{center}
			\caption{Projective limit: $f^{-1}b$}
			\label{fig:preimage}
		\end{minipage}
	}
	\subfigure{
	\begin{minipage}[t]{0.4\linewidth}	
		\begin{center}
			\tikzset{every picture/.style={line width=0.75pt}} 

\begin{tikzpicture}[x=0.75pt,y=0.75pt,yscale=-1,xscale=1]
	
	\draw [color={rgb, 255:red, 0; green, 0; blue, 255 }  ,draw opacity=1 ]   (138.4,204.48) -- (151.48,229.7) ;
	\draw [shift={(152.4,231.48)}, rotate = 242.59] [color={rgb, 255:red, 0; green, 0; blue, 255 }  ,draw opacity=1 ][line width=0.75]    (10.93,-3.29) .. controls (6.95,-1.4) and (3.31,-0.3) .. (0,0) .. controls (3.31,0.3) and (6.95,1.4) .. (10.93,3.29)   ;
	\draw [color={rgb, 255:red, 0; green, 0; blue, 255 }  ,draw opacity=1 ]   (211.6,199.75) -- (189.58,229.87) ;
	\draw [shift={(188.4,231.48)}, rotate = 306.17] [color={rgb, 255:red, 0; green, 0; blue, 255 }  ,draw opacity=1 ][line width=0.75]    (10.93,-3.29) .. controls (6.95,-1.4) and (3.31,-0.3) .. (0,0) .. controls (3.31,0.3) and (6.95,1.4) .. (10.93,3.29)   ;
	\draw [color={rgb, 255:red, 0; green, 0; blue, 0 }  ,draw opacity=0.4 ]   (194.4,201.48) -- (185.41,216.76) ;
	\draw [shift={(184.4,218.48)}, rotate = 300.47] [color={rgb, 255:red, 0; green, 0; blue, 0 }  ,draw opacity=0.4 ][line width=0.75]    (10.93,-3.29) .. controls (6.95,-1.4) and (3.31,-0.3) .. (0,0) .. controls (3.31,0.3) and (6.95,1.4) .. (10.93,3.29)   ;
	\draw [color={rgb, 255:red, 0; green, 0; blue, 0 }  ,draw opacity=0.4 ]   (150.4,201.48) -- (156.64,216.63) ;
	\draw [shift={(157.4,218.48)}, rotate = 247.62] [color={rgb, 255:red, 0; green, 0; blue, 0 }  ,draw opacity=0.4 ][line width=0.75]    (10.93,-3.29) .. controls (6.95,-1.4) and (3.31,-0.3) .. (0,0) .. controls (3.31,0.3) and (6.95,1.4) .. (10.93,3.29)   ;
	
	\draw (154,241) node [anchor=north west][inner sep=0.75pt]  [color={rgb, 255:red, 0; green, 0; blue, 255 }  ,opacity=1 ] [align=left] {X$\displaystyle +$Y};
	\draw (157.4,221.48) node [anchor=north west][inner sep=0.75pt]  [color={rgb, 255:red, 0; green, 0; blue, 0 }  ,opacity=0.4 ] [align=left] {x \ \ y};
	\draw (127,181) node [anchor=north west][inner sep=0.75pt]  [color={rgb, 255:red, 255; green, 0; blue, 0 }  ,opacity=1 ] [align=left] {X};
	\draw (211,181) node [anchor=north west][inner sep=0.75pt]  [color={rgb, 255:red, 255; green, 0; blue, 0 }  ,opacity=1 ] [align=left] {Y};
	\draw (144,187.5) node [anchor=north west][inner sep=0.75pt]  [color={rgb, 255:red, 0; green, 0; blue, 0 }  ,opacity=0.4 ] [align=left] {x};
	\draw (192,186.5) node [anchor=north west][inner sep=0.75pt]  [color={rgb, 255:red, 0; green, 0; blue, 0 }  ,opacity=0.4 ] [align=left] {y};

\end{tikzpicture}
		\end{center}
		\caption{Inductive limit: $X+Y$}
		\label{fig:sum}
	\end{minipage}
}
	\subfigure{
	\begin{minipage}[t]{0.4\linewidth}	
		\begin{center}
			\tikzset{every picture/.style={line width=0.75pt}} 

\begin{tikzpicture}[x=0.75pt,y=0.75pt,yscale=-1,xscale=1]
	
	\draw [color={rgb, 255:red, 255; green, 0; blue, 0 }  ,draw opacity=1 ]   (138.07,177.67) -- (180.07,178.3) ;
	\draw [shift={(182.07,178.33)}, rotate = 180.87] [color={rgb, 255:red, 255; green, 0; blue, 0 }  ,draw opacity=1 ][line width=0.75]    (10.93,-3.29) .. controls (6.95,-1.4) and (3.31,-0.3) .. (0,0) .. controls (3.31,0.3) and (6.95,1.4) .. (10.93,3.29)   ;
	\draw [color={rgb, 255:red, 255; green, 0; blue, 0 }  ,draw opacity=1 ]   (138.73,186.33) -- (180.73,186.97) ;
	\draw [shift={(182.73,187)}, rotate = 180.87] [color={rgb, 255:red, 255; green, 0; blue, 0 }  ,draw opacity=1 ][line width=0.75]    (10.93,-3.29) .. controls (6.95,-1.4) and (3.31,-0.3) .. (0,0) .. controls (3.31,0.3) and (6.95,1.4) .. (10.93,3.29)   ;
	\draw [color={rgb, 255:red, 0; green, 0; blue, 255 }  ,draw opacity=1 ]   (192.4,196) -- (192.7,215.33) ;
	\draw [shift={(192.73,217.33)}, rotate = 269.1] [color={rgb, 255:red, 0; green, 0; blue, 255 }  ,draw opacity=1 ][line width=0.75]    (10.93,-3.29) .. controls (6.95,-1.4) and (3.31,-0.3) .. (0,0) .. controls (3.31,0.3) and (6.95,1.4) .. (10.93,3.29)   ;
	
	\draw (122,176.33) node [anchor=north west][inner sep=0.75pt]  [color={rgb, 255:red, 255; green, 0; blue, 0 }  ,opacity=1 ] [align=left] {X};
	\draw (187.07,177.33) node [anchor=north west][inner sep=0.75pt]  [color={rgb, 255:red, 255; green, 0; blue, 0 }  ,opacity=1 ] [align=left] {Y};
	\draw (152,159.67) node [anchor=north west][inner sep=0.75pt]  [color={rgb, 255:red, 255; green, 0; blue, 0 }  ,opacity=1 ] [align=left] {f};
	\draw (150.67,192.33) node [anchor=north west][inner sep=0.75pt]  [color={rgb, 255:red, 255; green, 0; blue, 0 }  ,opacity=1 ] [align=left] {g};
	\draw (175.33,224) node [anchor=north west][inner sep=0.75pt]  [color={rgb, 255:red, 0; green, 0; blue, 255 }  ,opacity=1 ] [align=left] {Y/$\displaystyle \sim $};
	\draw (201.67,202.33) node [anchor=north west][inner sep=0.75pt]  [color={rgb, 255:red, 0; green, 0; blue, 255 }  ,opacity=1 ] [align=left] {p};

\end{tikzpicture}
		\end{center}
		\caption{Inductive limit: quotient set}
		\label{fig:equalizer}
	\end{minipage}
}
\end{figure}

When the functor $\rlim\alpha$ (resp.~$\llim\beta$) is representable, we use the same notation  $\rlim\alpha\in \cat$ (resp.~$\llim\beta\in\cat$) to denote its representative. 
If for every diagram from $I$ (or $I^\op$) to $\cat$, $\rlim \alpha$ (or $\llim \beta$) is representable, we say $\cat$ admits inductive (or projective) limits indexed by $I$. If $\cat$ admits inductive (or projective) limits indexed by all small categories, we say $\cat$ admits small projective (or inductive) limits.
Below we provide a few examples of the limits, 
where we use {\color{red} red} to denote the diagram,
and {\color{blue} blue} to denote the limit of the diagram.

\textbf{Product $X\times Y$.} 
Given $X, Y\in \cat$, $X\times Y$ can be seen as the limit of two different diagrams. 
\begin{enumerate}
	\item It is the projective limit of the functor $\beta: \iop\rightarrow \cat$, where $\iop$ is a discrete category of two objects $\{I_1,I_2\}$, and $\beta(I_1)=X$, $\beta(I_2)=Y$.
	It simply means we will map any $(x,y)\in X\times Y$ to $x\in X$ and $y\in Y$.  See Figure~\ref{fig:product1} left. 
	\item It is also the projective limit of the pullback functor $\beta: \iop\rightarrow \cat$, where $\iop$ is a category of three objects $\{I_1,I_2, I_3\}$, and $\beta(I_1)=X$, $\beta(I_2)=Y, \beta(I_3)=\{\text{pt}\}$. Moreover, there are two non-identity morphisms: $m_1: I_1\rightarrow I_3$, $m_2:I_2\rightarrow I_3$. Denote $f\triangleq \beta(m_1), g\triangleq \beta(m_2)$. Since $\beta(I_3)$ is the single point set, we know both $f$ and $g$ are the collapse morphism that maps everything to the same point. We use the $X\times_* Y$ to represent the projective limit of this diagram. By definition, we know $X\times_* Y=\{
	(x,y)\in X\times Y| f(x)=g(y)=\{\text{pt}\}
	\}=X\times Y$. See Figure~\ref{fig:product1} right.
\end{enumerate}

\textbf{Preimage.} It is the projective limit of the equalizer functor $\beta:\iop\rightarrow \cat$, where $\iop$ is a category of two objects $\{I_1, I_2\}$, with two morphisms $m_1, m_2$ in between. Denote $\beta(I_1)=X, \beta(I_2)=Y$, and $\beta(m_1)=f, \beta(m_2)=g$, both are morphisms from $X$ to $Y$. If $g$ is the morphism that sends everything to $b$,  the projective limit of $\beta$ is the set $\{(x,y)\in X\times Y| f(x)=g(x)=y=b
\}$, in other words, this set is isomorphic to the set $f^{-1}b\in X$. See Figure~\ref{fig:preimage}.

\textbf{Sum $X+Y$.} Given $X, Y\in \cat$, $X+Y$ can be seen as the inductive limit of the function $\alpha: I\rightarrow \cat$, where $I$ is a discrete category of two objects $\{I_1, I_2\}$, and $\alpha(I_1)=X, \alpha(I_2)=Y$. It simply means we will combine the elements in $X$ and $Y$ together, to form a bigger object $X+Y$. 
See Figure~\ref{fig:sum}.

\textbf{Quotient.} It is the inductive limit of the coequalizer functor $\alpha:I\rightarrow \cat$, where $I$ is a category of two objefts $\{I_1, I_2\}$, with two morphisms $m_1, m_2$ in between. Denote $\alpha(I_1)=X, \alpha(I_2)=Y$, and $\alpha(m_1)=f, \alpha(m_2)=g$, both are morphisms from $X$ to $Y$. If we define the equivalence relation as $f(x)\sim g(x)$ for all $x\in X$, 
the inductive limit of $\alpha$ becomes 
the largest set of the form $\{
p(f(x)) ~|~ p(f(x))=p(g(x)) \text{~ for all~} x\in X
\}$ for some $p$, 
which is $Y/\sim$. See Figure~\ref{fig:equalizer}.

\subsection{Ind-lim and Pro-lim}
The limits in the category of presheaves ($\catw$ and $\catv$) have special names, defined below. 
\begin{mydef}[Ind-lim and Pro-lim]
Given $\alpha: I \rightarrow \catw, \beta: I^\op \rightarrow \catv$, 
we denote $\illim \alpha$ as the inductive limit of $\alpha$, 
and $\irlim \beta$ as the projective limit of $\beta$.  
\end{mydef}

We add quotation marks for these limits, because we want to overload the notation for the diagrams valued in $\cat$. Specifically, for $\alpha: I\rightarrow \cat$, we define $\irlim \alpha\triangleq \irlim (\hc\circ \alpha)$. For $\beta: I^\op\rightarrow \cat$, we define $\illim \beta \triangleq \illim (\kc \circ \beta)$. In other words, $\irlim$ and $\illim$ have two kinds of inputs, and the quotation marks means the diagram will be lifted to $\catw$ and $\catv$ if necessary. 

\begin{mylem}[Corollary 2.4.3 in \citet{kashiwara2006categories}]
	\label{lem:representable}
	Both $\catw$ and $\catv$ admit small inductive and small projective limits.  	
\end{mylem}

\section{Concept}
\label{sec:concept}
We first give the formal definition of a concept, as well as a valid decomposition of a given concept. 

\begin{mydef}[Concept]
\label{def:concept}
Given a category $\cat$, a concept $A$ is a presheaf in $\catw$ or $\catv$. 
\end{mydef}

\begin{mydef}[Valid decomposition under $T$]
	\label{def:valid-under-T}
Given a concept $A$, a decomposition $\mathcal{D_A}$ is valid under task $T$ for $A$, if applying $T$ to $A$ is equivalent to applying $T$ to each subconcept in $\mathcal{D_A}$ individually and then combine the results together with (possibly recursive) application of projective or inductive limits.
\end{mydef}

Definition~\ref{def:valid} can be seen as the weaker version of saying $T$ commutes with the decomposition operator. For example, if we use $\rlim \alpha$ to represent the decomposition of $A$, $T$ commutes with $\rlim \alpha$ means $T(\rlim \alpha)\simeq \rlim T(\alpha)$, but 
this decomposition is still valid if $T(\rlim \alpha)\simeq \llim T(\alpha)$. More generally, the decomposition can be a hierarchical tree, where the combination step recursively merge the leaf nodes to their fathers, see Section~\ref{sec:example} for example. 

We consider two kinds of tasks, the morphisms $\hom_\cat(\cdot, A)$ or $\hom_\cat(A, \cdot)$, and the functors $F:\cat\rightarrow \cat'$. 
As a result, a valid decomposition  for $A$ maintains all the relationships and functionalities invariant.

\begin{mydef}[Valid decomposition]
	\label{def:valid}
	Given a concept $A$, when a decomposition $\mathcal{D_A}$ is valid for $A$ under all possible morphisms and functors, we say it is valid for $A$. 	
\end{mydef}

\subsection{Limits as representations}
Definition~\ref{def:valid} naturally leads to the projective and inductive limits in $\cat$. Indeed, we have the following two lemmas. 

\begin{mylem}[Limits for morphism tasks]
\label{lem:limit-mor}
Consider $\alpha: I\rightarrow \cat, \beta: I^\op \rightarrow \cat$ with small $I$. If $\rlim\alpha$ and $\llim \beta$ are representable, respectively, we have
\[
\hom_\cat(\rlim \alpha, X)\simeq \llim \hom_\cat(\alpha, X)
\]	
\[
\hom_\cat(X, \llim \beta)\simeq \llim \hom_\cat(X, \beta)
\]
\end{mylem}

Lemma~\ref{lem:limit-mor} easily follows by Definition~\ref{def:limits}.

\begin{mylem}[Limits for functor tasks, Proposition 2.1.10 in \citet{kashiwara2006categories}]
\label{lem:limit-functor}
Let $F:\cat\rightarrow \cat'$ be a functor. Assume that:
\begin{itemize}
	\item $F$ admits a left adjoint $G:\cat'\rightarrow \cat$, 
	\item $\cat$ admits projective limits indexed by a small category $I$. 
\end{itemize}
Then $F$ commutes with projective limits indexed by $I$, that is, the natural morphism $F(\llim \beta)\rightarrow \llim F(\beta)$ is an isomorphism for any $\beta:I^\op \rightarrow \cat$. 
Similar result holds for inductive limits. If $\cat$ admits inductive limits indexed by $I$, and $F$ admits a right adjoint, then $F$ commutes with such limits. 
\end{mylem}

Lemma~\ref{lem:limit-mor} and Lemma~\ref{lem:limit-functor} indicates that both projective and inductive limits can be seen as valid decompositions for morphism and functor tasks. However, they are not perfect for the following reasons: 

\begin{enumerate}
	\item \textbf{Order matters} in Lemma~\ref{lem:limit-mor}. The inductive limit operator $\rlim$commutes with $\hom_\cat$ only when it appears in domain, but not in codomain (e.g., $\hom_\cat(X, \rlim \alpha)$). Similarly,  the projective limit operator $\llim$commutes  with $\hom_\cat$ only when it appears in codomain, but not in domain (e.g., $\hom_\cat(\llim \beta, X)$).
	
	\item The concept has to be \textbf{representable} for both lemmas. Otherwise, the limits are not well defined in $\cat$. 
	
	\item The functor $F$ must \textbf{admit a left adjoint} in Lemma~\ref{lem:limit-functor}. 	
\end{enumerate}

\subsection{Ind-lim and Pro-lim as representations}
It turns out that the previous problems can be easily solved in the category of presheaves. Moreover, from the deep learning perspective, it is more convenient to work with the feature space, instead of the input space. As discussed in~\citet{yuan2022power}, the feature space of the neural network can be seen as the empirical representations for presheaves.

Extending Lemma 2 to presheaves, we immediately have the following lemma:
\begin{mylem}
	\label{lem:ind-mor}
If $I$ is small, consider $\alpha: I\rightarrow \catw, \beta: I^\op \rightarrow \catv$. 
For $A\in \catw, B\in \catv$,
\[
\hom_\catw(\irlim \alpha, A)\simeq \llim \hom_\catw(\alpha, A)
\]	
\[
\hom_\catv(B, \illim \beta)\simeq \llim \hom_\catv(B, \beta)
\]
\end{mylem}
Notice that $A$ and $B$ are presheaves, therefore they can be treated as functors in $\cat^\op\rightarrow \set$ or $\cat^\op \rightarrow \set^\op$. When $\alpha$ is representable, denote $X\in\cat$ as the representative, we have $\hom_\catw(X, A)\simeq A(X)$ by Yoneda lemma. Interestingly, 
unlike Lemma~\ref{lem:limit-mor}, we have guarantees for the other side of the morphisms:
\begin{mylem}[(2.6.1-2) in \citet{kashiwara2006categories}]
\label{lem:ind-mor2}
If $I$ is small, consider $\alpha: I\rightarrow \catw, \beta: I^\op \rightarrow \catv$. 
For $X\in \cat$,
\[
\hom_\catw(X,\irlim \alpha)\simeq \rlim \hom_\catw(X, \alpha)
\]
\[
\hom_\catv(\illim \beta, X)\simeq \rlim \hom_\catv(\beta,X)
\]
\end{mylem}

Lemma~\ref{lem:ind-mor} and Lemma~\ref{lem:ind-mor2} indicate that both ind-lims and pro-lims can be seen as valid decompositions for morphism and $\set$-valued functor tasks. They are better than the projective and inductive limits in $\cat$ for the following reasons:

\begin{enumerate}
	\item Order does not matter for the morphisms. Indeed, combining the two lemmas, we can see for any given object $X\in \cat$, $X$ may appear in both domain and codomain in $\hom(\cdot, \cdot)$. 
	\item The ind-lims and Pro-lims are always representable, due to Lemma~\ref{lem:representable}.
	\item There is no restrictions on the functor $F$ in Lemma~\ref{lem:ind-mor}. $F$ can be any functor in $\catw$ or $\catv$. 
\end{enumerate}

For the more general functor $F:\cat\rightarrow\cat'$, we have the following two lemmas. The first lemma says, as long as $\irlim \alpha$ is representable,  the decomposition is valid under any $F$.

\begin{mylem}[Proposition 2.6.4 in \citet{kashiwara2006categories}]
	\label{lem:lim-representable}
	Let $I$ be a small category and $\alpha: I\rightarrow \cat$ a diagram. Assume that $\irlim \alpha\in \catw$ is represented by  $X\in \cat$, then for any functor $F:\cat\rightarrow \cat', \llim (F\circ \alpha)\simeq F(X)$. Similar results hold for $\illim \beta$ with $\beta:I^\op\rightarrow \cat$ as well.
\end{mylem}
Lemma~\ref{lem:lim-representable} requires $\irlim \alpha$ to be representable, and works in $\cat$ instead of the category of presheaves. This is not convenient because foundation models work in the feature space, especially for prompt tuning or fine tuning. Therefore the next lemma is more relevant. 

\begin{mylem}[Proposition 2.7.1 in \citet{kashiwara2006categories}]
	\label{lem:yoneda-extension}
Assume $\cat$ is small and $\cat'$ admits small inductive limits. 
Given $F:\cat \rightarrow \cat'$, there exists $\tilde{F}:\catw\rightarrow \cat'$, such that $\tilde{F}\circ \hc\simeq F$ and $\tilde{F}$ commutes with small inductive limits: $\tilde{F}(\irlim \alpha)\simeq \rlim(F\circ \alpha)$ for any $\alpha:I\rightarrow \cat$ with $I$ small.
Similar results hold for $\catv$and $\illim \beta$ as well.
\end{mylem}
Lemma~\ref{lem:yoneda-extension} directly works in the category of presheaves, and shows that the decompositions are valid under any functor $F$. 
Therefore,  it becomes clear that Ind-lims and Pro-lims are better representations for concepts. We have the following definitions to distinguish basic and complex concepts. 

\begin{mydef}[Basic concept]
A concept $A$ is basic, if $A$ is representable and there is no 
$\alpha: I\rightarrow \cat$ such that $\irlim \alpha\simeq A$,  and no $\beta: I^\op \rightarrow \cat$ such that $\illim \beta\simeq A$. 
\end{mydef}

We remark that $\alpha, \beta$ are valued in $\cat$ instead of $\catw$ or $\catv$. This is to make hierarchical decompositions possible, as we will discuss in Section~\ref{sec:hierarchical}.

\begin{mydef}[Complex concept]
A concept $A$ is complex, if there exists $\alpha: I\rightarrow \cat$ such that $\irlim \alpha\simeq A$,  or $\beta: I^\op \rightarrow \cat$ such that $\illim \beta\simeq A$. 
\end{mydef}

Interestingly, if a presheaf is not representable, and not isomorphic to any limits, it is neither a basic concept nor a complex concept. We do not consider these cases in our paper. Therefore, when we talk about a concept below, we assume it is either basic or complex.

\subsection{Concept transfer}
If the concepts are defined as limits, can we transfer concepts from one category to another? This directly relates to how functors preserve or reflect limits. 
\begin{mydef}[Preservation and reflection of limits]
Given $\alpha:I\rightarrow \cat$ with small $I$, $F:\cat\rightarrow \cat'$, we say
\begin{itemize}
	\item $F$ preserves the limits $\rlim \alpha$, if whenever $\rlim\alpha$ exists,  $
	F(\rlim \alpha)\simeq \rlim F(\alpha)$. 
	\item $F$ reflects the limits, 
if whenever	$\rlim F(\alpha)\simeq F(X)$, 
we know $\rlim \alpha$ exists and $\rlim \alpha\simeq X$. 
\end{itemize}
Similar definitions hold for the projective limits as well.
\end{mydef}

``$F$ preserves the limits'' is equivalent to 
``$F$ commutes with the limits''. Therefore by Lemma~\ref{lem:yoneda-extension}, $\tilde F$ preserves limits. If we treat the limits as concepts, there are many interesting implications. For example, consider the CLIP model~\citep{radford2021learning}, which learns a functor that connects the text category and image category~\citep{yuan2022power}. The current implementation of CLIP relies on the InfoNCE loss, which constructs a similarity graph on the image and text objects. If one injects more complicated semantics during training, so that both categories can have deeper concepts, by Lemma~\ref{lem:yoneda-extension}, we will be able to generate more semantically meaningful images with a limit-preserving functor, and generate 
more comprehensive text descriptions with a limit-reflecting functor. 

\subsection{Definitions, examples and analogies}
Intuitively, the projective limit and inductive limit can be seen as two ways of learning new concepts: learn by definition and learn by examples. Specifically, the projective limit captures the underlying elements supporting the concept, and the inductive limit captures the similarities shared by the underlying examples in the representation.

Making analogies is another typical way of learning new concepts. 
It can be seen as making a transformation between the limits of two  diagrams $\alpha:I\rightarrow \cat$ and $\beta:J\rightarrow \cat'$. If we are very familiar with one limit (concept), with the help of this transformation we may quickly grasp the other concept as well. Therefore, the similarity between two concepts can be modeled as the similarity between the two diagrams. However, this transformation does not decompose a given concept, so is not counted as a succinct representation.

Being able to make analogies is important.
For example, human beings may ask chatGPT dangerous questions like ``how to destroy the world?'' We certainly do not want AI to answer this question. 
The researchers tried hard to filter these dangerous questions,
but empirically the users are creative enough to generate semantically different but inherently similar questions to bypass the filter. 
To solve this problem, we may prevent AI from answering any questions containing concepts that are similar to the dangerous ones. On the other hand, if the question contains concepts that are very different to any dangerous concepts, it shall be safe to speak out the answer.

\section{Algorithms}
\subsection{Hierarchical decomposition}
\label{sec:hierarchical}
We have seen that 
a complex concept can be decomposed using Ind-lims and Pro-lims. How can we do it recursively? Notice that $\hc$ commutes with projective limits but not inductive limits, and $\kc$ commutes with inductive limits but not projective limits. If we want to include both kinds of limits in our decomposition, we cannot stay inside the same category. Otherwise the presheaves we operate will not have a representative in $\cat$, therefore they cannot be easily understood by human beings. 

To solve this problem, we only consider $\irlim \alpha$ and $\illim \beta$ for  $\alpha:I\rightarrow \cat, \beta:I^\op \rightarrow \cat$. The diagrams are valued in $\cat$ instead of $\catw$ or $\catv$ to make the hierarchical decomposition feasible.  Specifically, after each decomposition, we obtain a diagram on the objects representable in $\cat$. 
If further decompositions are needed, 
we may use $\hc$ (or $\kc$) to lift the object to $\catw$ (or $\catv$), depending on whether it is a projective or inductive limit. 
Notice that $\hc$ commutes with projective limits, so we can directly decompose it in $\catw$ (same for $\kc$). Using this trick, recursive decomposition is possible, and we may jump between $\catw$ and $\catv$ during the decomposition. See Algorithm~\ref{alg:decompose} for details, which assumes the existence of a limit extractor defined below. 

\begin{mydef}[Limit extractor]
	\label{def:limit-extractor}
Given a foundation model $f$, a concept $A$, a limit extractor $g$ outputs 
a limit $g(f,A)$ that contains a limit operator and a diagram, i.e., $\irlim \alpha$ for $\alpha:I\rightarrow \cat$ or $\illim\beta$ for $\beta: I^\op\rightarrow \cat$.  
\end{mydef}

In Definition~\ref{def:limit-extractor}, we did not assume $\irlim g(f,A)\simeq A$ or $\illim g(f,A)\simeq A$, because $g$ is a function that we have to implement. Ideally, it shall output a diagram whose limit is equivalent to $A$, but empirically it may make mistakes. It is possible that there are different limits for a concept, and here we assume that $g$ picks a random one, or follows some specific rules.

\begin{algorithm}
	\caption{\dec}
	\label{alg:decompose}

\begin{algorithmic}
	\STATE {\bfseries Input:} foundation model $f$, concept $A$ in $\catw$ or $\catv$,  limit extractor $g$
	\STATE Let $\mathcal{L_A}=g(f,A)$
	\IF {$\mathcal{L_A}$ not null}
	\STATE Let $\mathcal{D_A}=\{\}$
	\FOR{$A_i$ in $\mathcal{L_A}$}
	   \STATE Let $L_{A_i}=g(f,A_i)$
	   \IF {$L_{A_i}$ not null}
		  \IF {$L_{A_i}$ is projective limit}
			   \STATE Let $\mathcal{D_A}=\mathcal{D_A}+$\dec($f, \hc(A_i), g$)
		  \ELSE
			   \STATE Let $\mathcal{D_A}=\mathcal{D_A}+$\dec($f, \kc(A_i), g$)
		  \ENDIF
	   \ELSE
	      \STATE Let $\mathcal{D_A}=\mathcal{D_A}+{A_i}$
	   \ENDIF
    \ENDFOR  
    \STATE return $\mathcal{D_A} $
	\ELSE
	\STATE return $A$
	\ENDIF
\end{algorithmic}
\end{algorithm}

\subsection{Decomposition example}
\label{sec:example}
Consider we want to decompose the concept ``dynamic programming'', which may have both (one or more) inductive and projective limits. As the projective limit, it means ``an algorithm that breaks down the original problem into sub-problems, and solve the sub-problems recursively'', represented by a diagram.  As the inductive limit, it may contain different types of dynamic programming, e.g., DP on one-dimensional array, DP on two-dimensional array, DP on higher-dimensional array, DP on graphs, etc. 

Most of the sub-concepts can be further decomposed. For example, The concept of ``sub-problem recursion'' may have a projective limit that describes how to recursively run a sub-problem until the boundary case. In this description, there may exists a sub-concept like ``for-loop'', which is similar to another concept ``while-loop'' by making a diagram analogy. The concept of ``DP on one-dimensional array'' may have a projective limit describing how to 
transform the problem into the one-dimensional array,
how to store the optimal state, how to define the recursion, etc. 

With large amount of training data, it is possible that the diagrams used by the foundation models will contain much more details, but the overall structure of the hierarchical decomposition should similar.

\subsection{Main Results}

If $\mathcal{D_A}$ is generated using Algorithm~\ref{alg:decompose}, 
it is not necessarily valid because the limit extractor may make mistakes. 
Our main theorem says that, being able to recursively generating the correct limits of each concept in $\mathcal{D_A}$ is equivalent to precise comprehension of the concept.

\begin{mythm}[Main theorem]
\label{thm:main}
	A foundation model $f$ with the limit extractor $g$, precisely understands $A$, if every non-leaf node in $\mathcal{D_A}\triangleq \dec(f,A,g)$ is isomorphic to the limit extraction given by~$g$. 
\end{mythm}

Applying Theorem~\ref{thm:main} for learning $f$, we  have the following corollary. See Algorithm~\ref{alg:learning} for details, where $T(\mathcal{L_A})$ means applying~$T$ to each subconcept in $\mathcal{L_A}$ and then combine the results together.

\begin{algorithm}
	\caption{Concept learning}
	\label{alg:learning}
	
	\begin{algorithmic}
		\STATE {\bfseries Input:} foundation model $f$, a  concept $A$ in $\catw$ or $\catv$,  limit extractor $g$		
		\FOR{$i=1$ to $m$}
		\STATE Sample a task $T$ for $A$
		\STATE Let $\mathcal{L_A}=g(f,A)$
		\STATE Let $\ell(T,A, \mathcal{L_A})=
		d(T(A), T(\mathcal{L_A}))$ for a similarity function $d$ 
		\STATE Run backpropagation on $\ell(T,A, \mathcal{L_A})$ to optimize $f$ and $g$
		\ENDFOR
	\end{algorithmic}
\end{algorithm}

\begin{mycor}
\label{cor:learning}
To learn $f$ that precisely understand the concepts, one should learn a limit extractor $g$ such that given any concept $A$, for all possible task $T$, $g(f,A)$ is valid under $T$. 
\end{mycor}

Algorithm~\ref{alg:learning} can be easily modified to fix a misconception, or learn a new concept. Specifically, if we know the correct representation for a concept $A$, denoted as $\mathcal{L_A}^*$, we may simply run supervised learning for $f$ and $g$, so that $\mathcal{L_A}^*=g(f,A)$. 

Apply Theorem~\ref{thm:main} for verifying a concept $A$ with $f$, we 
have the following corollary. 

\begin{algorithm}
	\caption{Concept verification with tasks}
	\label{alg:verify-1}	
	\begin{algorithmic}
			\STATE {\bfseries Input:} foundation model $f$, a  concept $A$ in $\catw$ or $\catv$,  limit extractor $g$		
	\STATE Let $\mathcal{L_A}=g(f,A)$
	\IF {$\mathcal{L_A}$ not null}
		\FOR{$i=1$ to $m$}
		\STATE Sample a task $T$ for $A$
		\STATE Let $\ell(T,A, \mathcal{L_A})=
		d(T(A), T(\mathcal{L_A}))$ for a similarity function $d$
		\STATE Return failure if $\ell(T,A, \mathcal{L_A})\geq \epsilon$ 
		\ENDFOR
	\ENDIF 	
	\STATE Return success
	\end{algorithmic}
\end{algorithm}

\begin{mycor}
\label{cor:verify}
To verify that $f$ has precisely understood a concept $A$ with limit extractor $g$, one should
use $g$ to produce a decomposition $\mathcal{D_A}$, and verify 
the decomposition of every non-leaf node in $\mathcal{D_A}$ is a projective or inductive limit. 
\end{mycor}

There are two kinds of algorithms for verifying a concept. 
Algorithm~\ref{alg:verify-1} is similar to Algorithm~\ref{alg:learning} as it verifies concepts by sampling various tasks. Both Algorithm~\ref{alg:learning} and Algorithm~\ref{alg:verify-1} can be extended to the hierarchical decomposition by calling Algorithm~\ref{alg:decompose}, which we omit here.

\begin{algorithm}
	\caption{Concept verification with verifier}
	\label{alg:verify-2}	
	\begin{algorithmic}
		\STATE {\bfseries Input:} foundation model $f$, a  concept $A$ in $\catw$ or $\catv$,  limit extractor $g$, a limit verifier $v$
		\STATE let $\mathcal{D_A}=$\dec($f,A,g$)
		\FOR{non-leaf concept $B$ in $\mathcal{D_A}$}
		\STATE Let $\mathcal{L_B}=g(f,B)$
		\STATE Return failure if $v(\mathcal{L_B}, B)=0$ 
		\ENDFOR
		\STATE Return success
	\end{algorithmic}
\end{algorithm}

Algorithm~\ref{alg:verify-2} assumes the existence of a limit verifier, defined below. 

\begin{mydef}[Limit verifier]
Given a foundation model $f$, a concept $A$, a limit extractor $g$, a limit verifier $v$ is a function that verifies $g(f,A)$ by computing $v(g(f,A),A)\triangleq  I_{g(f,A)\simeq A}$. 
\end{mydef}

The limit verifier can be an existing well trained foundation model, which serves as a teacher that teaches the current model $f$. It can also be a human-being, who verifies whether the limit $g(f,A)$ is true (especially for math concepts). It is worth pointing out that we assume that $f$ precisely understands the leaf nodes, just like how chatGPT understands basic concepts by training with billions of tasks. The correctness of Algorithm~\ref{alg:verify-2} easily follows by this assumption and Theorem~\ref{thm:main}.

\section{Discussions}
Almost all the branches of modern mathematics  can be described using category theory, and complicated math concepts are defined using limits from the objects in various categories. A theorem (or lemma) can be seen as an isomorphism between two concepts. Therefore, it is conceivable that AI will be able to solve math problems after accurately learning the concepts using our algorithms. 

There are numerous other applications as well. For example, it can be used for programming. Each subroutine of the program has certain functionality, which corresponds to a concept. If we can learn a functor that transfer the concept from the natural language category to the program category, AI will be able to write code automatically with language hints. Although chatGPT has demonstrated amazing power on generating code, it will be able to think deeper after learning the concepts.

Moreover, our characterization for the concepts as presheaves easily explains why the existing foundation models are super good at creating plausible and non-existent stories on well-known characters. For example, if we ask chatGPT to ``tell what Odysseus will say to Confucius'', it will generate a long conversation based on their respective beliefs and values. Such conversation certainly never existed before in the training data, but the model understands Odysseus and Confucius by learning other related materials, and encodes them into high dimensional vectors to approximate their corresponding functors (i.e. presheaves) that can predict their behaviors for all possible tasks, including a hypothetical conversation. In other words, when the model sees words like ``Odysseus'' or ``Confucius'', it sees not only the words, but also all the relationships and properties, just like our human beings.

\bibliographystyle{apalike}
\bibliography{paper}  






\end{document}